\documentclass[preprint]{article}
\usepackage{style/neurips_2024}

\usepackage[utf8]{inputenc}
\usepackage[T1]{fontenc}
\usepackage{enumitem}
\usepackage{graphicx}
\usepackage{pifont}
\usepackage{url}
\usepackage{booktabs}        
\usepackage{amsfonts}        
\usepackage{nicefrac}        
\usepackage{microtype}       
\usepackage{xcolor}          
\usepackage{graphicx}
\usepackage{subcaption}

\title{OceanAI: A Conversational Platform for Accurate, Transparent, Near-Real‑Time Oceanographic Insights}

\author{
  Bowen Chen \\
  North Carolina State University\\
  \texttt{bchen39@ncsu.edu}\\
  \And
  Jayesh Gajbhar \\
  North Carolina State University\\
  \texttt{jgajbha@ncsu.edu}\\
      \And
  Gregory Dusek \\
  NOAA \\
  \texttt{gregory.dusek@noaa.gov}\\
      \And
  Rob Redmon \\
  NOAA\\
  \texttt{rob.redmon@noaa.gov}\\
      \And
  Patrick Hogan \\
  NOAA\\
  \texttt{patrick.hogan@noaa.gov}\\
    \And
  Paul Liu \\
  North Carolina State University\\
  \texttt{jpliu@ncsu.edu}\\
    \And
  DelWayne Bohnenstiehl \\
  North Carolina State University\\
  \texttt{drbohnen@ncsu.edu}\\
    \And
  Dongkuan (DK) Xu \\
  North Carolina State University\\
  \texttt{dxu27@ncsu.edu}\\
    \And
  Ruoying He \\
  North Carolina State University\\
  \texttt{rhe@ncsu.edu}\\
}

\begin{document}
\maketitle

\begin{abstract}
Artificial intelligence is transforming the sciences, yet general conversational AI systems often generate unverified "hallucinations" undermining scientific rigor. We present \textbf{OceanAI}, a conversational platform that integrates the natural-language fluency of open-source large language models (LLMs) with real-time, parameterized access to authoritative oceanographic data streams hosted by the National Oceanic and Atmospheric Administration (NOAA). Each query—such as \textit{“What was Boston Harbor’s highest water level in 2024?”},triggers real-time API calls that identify parse and synthesize relevant datasets into
reproducible natural-language responses and data visualizations. In a blind comparison with three widely used AI chat-interface products, only OceanAI produced NOAA-sourced values with original data references; others either declined to answer or provided unsupported results. Designed for extensibility, OceanAI connects to multiple NOAA data products and variables, supporting applications in marine hazard forecasting, ecosystem assessment, and water-quality monitoring. By grounding outputs, verifiable observations, OceanAI advances transparency, reproducibility, and trust, offering a scalable framework for AI enabled decision support within the oceans. A public demonstration is available at \url{https://oceanai.ai4ocean.xyz}.
\end{abstract}
\vspace{-1em}
\begin{center}
\includegraphics[width=0.95\textwidth]{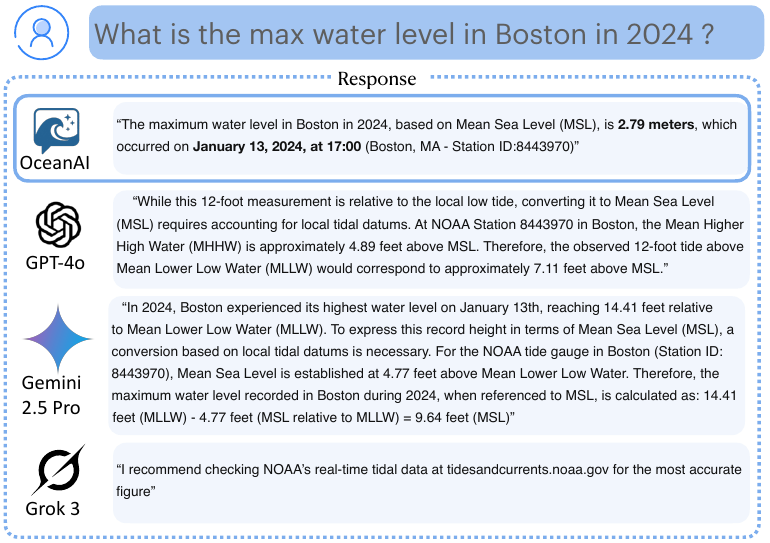}
\captionof{figure}{Comparison of model responses to “What is the maximum water level in Boston in 2024?”. OceanAI returns the correct NOAA-verified value (2.79 m MSL) with full metadata, while GPT-4o omits the value, Gemini-2.5 Pro miscalculates, and Grok-3 declines.}
\label{fig:abstract_comparison}
\end{center}
\vspace{-1em}


\section{Introduction}
\label{sec:introduction}

The ocean plays a pivotal role in regulating Earth's climate, supporting biodiversity, and sustaining coastal economies. Accurate understanding of oceanographic processes underpins climate forecasting, hazard preparedness, and sustainable resource management~\citep{hurrell2009unified,roemmich2019argo}. Over the past two decades, advances in satellite remote sensing, in-situ observation networks (e.g., Argo~\citep{Wong2020-rk}), and reanalysis programs(e.g., ERA5~\citep{ERA5} and GLORYS~\citep{GLORYS}) have dramatically expanded the availability of high-resolution ocean data. However, these datasets remain technically challenging to access and work
with due to their complex formats and spatiotemporal structure, creating barriers for non-expert users and overhead even for domain experts.

Existing tools for accessing oceanographic information fall broadly into three categories:general-purpose large language models (LLMs), domain-specific data servers, and hybrid geospatial analysis environments. General-purpose large language models (LLMs), such as GPT-4~\citep{openai2024gpt4technicalreport} and LLaMA~\citep{touvron2023llamaopenefficientfoundation}, offer intuitive natural language interaction but lack direct access to authoritative datasets. Without integration to structured data sources, these models cannot reliably execute parameterized scientific queries, and their outputs are prone to \textit{hallucination}~\cite{ji2023survey}, producing unverifiable or incorrect statements. Domain-specific platforms, such as the NOAA THREDDS Data Server\footnote{https://www.ncei.noaa.gov/thredds/catalog/catalog.html} and the NOAA Open Data Dissemination Program\footnote{https://www.noaa.gov/information-technology/open-data-dissemination}, provide structured, transparent access to observational and model data but require substantial technical expertise to navigate file hierarchies, parse specialized formats (e.g., NetCDF, GRIB), and perform downstream analysis. Hybrid geospatial analysis environments, such as Google Earth Engine~\citep{googleEarth}, provide access to global geospatial datasets—including some oceanographic layers—and support large-scale data processing and interactive visualization. However, they typically require users to write JavaScript or Python code and do not offer conversational interfaces or direct parameterized querying of authoritative ocean datasets.

To address these limitations, we present \textbf{OceanAI}, a tool-augmented conversational platform that combines the natural language fluency of modern LLMs with structured, callable access to authoritative oceanographic datasets. OceanAI overcomes the shortcomings of existing approaches through three key design strategies:
\begin{itemize}
    \item {\textbf{Direct data grounding: }}Queries are resolved into parameterized function calls to authoritative ocean datasets (e.g., NOAA) and extended to unstructured materials such as technical reports and scientific publications, ensuring responses integrate both verified data and contextual literature.
    
    \item \textbf{Automated data processing and visualization: }Retrieved datasets (e.g., NetCDF, GRIB) are transformed, analyzed, and visualized on the fly, lowering the technical barrier for users unfamiliar with specialized formats.
    \item \textbf{Transparent, up-to-date, and reproducible outputs: }Every response includes complete metadata on provenance, units, timestamps, and processing steps, enabling independent verification and reproduction of results. This design ensures that outputs reflect the most recent available observations from trusted providers.
\end{itemize}

\section{Ocean Background}
\subsection*{Sea Surface Temperature (SST) Data}

Sea surface temperature (SST) is widely used in earth sciences from ocean ecosystem conservation to weather forecasting. SST is usually measured by satellite sensors as well as in-situ platforms such as buoys and drifters. Prominent sources include NOAA's Coral Reef Watch and the Group for High Resolution Sea Surface Temperature(GHRSST), which offer daily global SST data at 5 km and finer resolution in NetCDF format. These files include spatial/temporal coordinates, quality flags, and uncertainty estimates, requiring coding expertise with scientific data packages such as xarray or netCDF4 for access~\citep{crw}.

SST is tightly linked to several important atmospheric events:

\begin{itemize}
  \item \textbf{El Niño-Southern Oscillation (ENSO):} Characterized by prolonged SST anomalies in the eastern tropical Pacific, ENSO events influence global atmospheric circulation and provoke extreme weather like droughts in South Asia and flooding in South America~\citep{elNinoReview,turn0search0}.
  \item \textbf{Monsoon variability:} Warmer SSTs in the Indian and Pacific Oceans enhance monsoon rainfall; cooler SSTs can suppress or delay monsoon onset, hindering agricultural production and affecting large populations~\citep{turn0search0,SINGHAI2024}.
  \item \textbf{Tropical cyclone intensity:} Storms feed on heat from the ocean surface. SST above 26.5 C significantly increases wind speed and rainfall of hurricanes / typhoons, causing significant damage to public safety and economy~\citep{Xu2016_SST_TC,nhess-18-795-2018}.
\end{itemize}

Identifying these phenomena requires computing SST anomalies from historical baselines, filtering spatial-temporal data, and applying threshold-based methods. However, the multi-dimensional nature and large volume of SST datasets present a technical barrier to users lacking programming or geospatial expertise.

\subsection*{Sea Level Data}

Sea level is monitored by tide gauges, such as by the NOAA Center\footnote{https://tidesandcurrents.noaa.gov/} for Operational Oceanographic Products and Services (CO-OPS) in the United States, and globally via the University of Hawaii Sea Level Center\footnote{https://uhslc.soest.hawaii.edu/}, which provide access to decades of  hourly and 6-minute water level observations.Distributed in CSV, NetCDF, or via API outputs (JSON/XML), these records are essential but require specialized technical skills and tools to process~\citep{moftakhari2015nuisance}.More recently, a high-resolution NOAA Coastal Ocean ReAnalysis (CORA)~\citep{CORA_data,CORA_report} enables a broader spatial representation of coastal sea level, however these data are similarly distributed in NetCDF and are large, with the entire data record exceeding 40TB. The large data size further complicates accessibility by requiring familiarity with cloud data storage and access.
For example,Sea level data are critical for understanding related natural hazards such as:
\begin{itemize}
  \item \textbf{High tide flooding:} Sea level rise has made ordinary high tides sufficient to flood low-lying urban areas (e.g., Boston, Norfolk), causing increasingly frequent “sunny-day” floods that disrupt traffic and infrastructure~\citep{li2021evolving,Hino2019-zf}.
  \item \textbf{Storm surge risk:} Elevated baseline sea levels combined with coastal storms dramatically worsen flood impact during hurricanes and typhoons~\citep{nhess-18-795-2018}.
  \item \textbf{Long-term rise:} Tide gauge data indicate global sea levels have increased by about 17cm from 1920-2020, contributing to more frequent and severe coastal flooding events~\citep{Sweet2022SLR}.
\end{itemize}


Extracting meaningful insights from sea-level records often requires understanding vertical datums, interpreting station or model metadata, and constructing time series from station logs or model outputs, which remain beyond the reach of many non-expert users.

\section{AI Background}
\label{sec:ai_background}

Recent advances in artificial intelligence have made large-scale language models (LLMs) accessible tools for scientific reasoning, data exploration, and user interaction. This section introduces the layered components that underpin modern natural language-based AI systems: 
(1) \textbf{large language models}, including their use of \textbf{embeddings} for semantic representation; 
(2) \textbf{retrieval-augmented generation} (RAG) for grounding outputs in external knowledge; and 
(3) \textbf{agentic RAG}, which extends RAG with dynamic tool selection and \textbf{function calling} to interface with computational infrastructures.

\subsection{Large Language Models (LLMs)}
Large language models such as GPT-4~\citep{openai2024gpt4technicalreport}, LLaMA~\citep{touvron2023llamaopenefficientfoundation}, and PaLM~\citep{chowdhery2022palm} are transformer-based neural networks~\citep{vaswani2023attentionneed} trained on large-scale text corpora to predict the next token in a sequence. This training paradigm enables them to generate coherent paragraphs, answer complex questions, and reason over natural language prompts.
Despite their impressive fluency, pretrained LLMs rely solely on patterns learned during training and have no inherent access to real-time or verifiable data sources.As a result, they may produce plausible-sounding but inaccurate outputs—a limitation known as \textit{hallucination}~\citep{ji2023survey}. This issue is especially problematic in scientific domains, where factual correctness and transparency are essential.

\subsubsection{Embeddings and Semantic Representations}
To support semantic reasoning and information retrieval, AI systems frequently use vector-based representations known as \textbf{embeddings}. An embedding is a numeric vector that encodes the meaning of a word, sentence, or document in a high-dimensional space. Similar concepts are mapped to geometrically close vectors, enabling systems to retrieve semantically related content even when surface forms differ.
Early embedding models like Word2Vec~\citep{mikolov2013efficient} and GloVe~\citep{pennington2014glove} assigned static vectors to words, whereas newer models such as BERT~\citep{devlin2019bert} and Sentence-BERT~\citep{reimers2019sentence} generate \textit{contextual embeddings}, allowing the same word to have different vector representations depending on its usage. These embeddings serve as an interface between human language and machine reasoning across tasks such as search, classification, and clustering.

\subsection{Retrieval-Augmented Generation (RAG)}
Retrieval-Augmented Generation (RAG) enhances the factuality of LLM outputs by integrating external document retrieval into the generation pipeline~\citep{lewis2020retrieval}. A RAG system typically follows a two-stage process:

\begin{itemize}
    \item The user query is embedded and compared against an external corpus to retrieve relevant documents~\citep{karpukhin2020dense}.
    \item The retrieved content is passed into the LLM's context window, allowing the model to generate responses grounded in real evidence~\citep{izacard2021leveraging}.
\end{itemize}

This architecture improves transparency and reduces hallucination by enabling responses conditioned on externally retrieved and verifiable information—making RAG particularly useful for scientific and high-stakes domains~\citep{gao2024retrieval}.

\subsection{Agentic RAG and Tool Use}
While classical RAG pipelines rely on retrieval alone, many real-world tasks—such as querying scientific datasets, generating visualizations, or running domain-specific simulations—require interaction with specialized computational tools. \textbf{Agentic RAG} extends the RAG paradigm by enabling models to plan, select, and invoke tools dynamically in response to user queries~\citep{masterman2024landscapeemergingaiagent}. This design allows the system to chain multiple steps, combining retrieval, computation, and synthesis into a coherent workflow.

\begin{figure}[!ht]
  \centering  \includegraphics[width=0.48\textwidth]{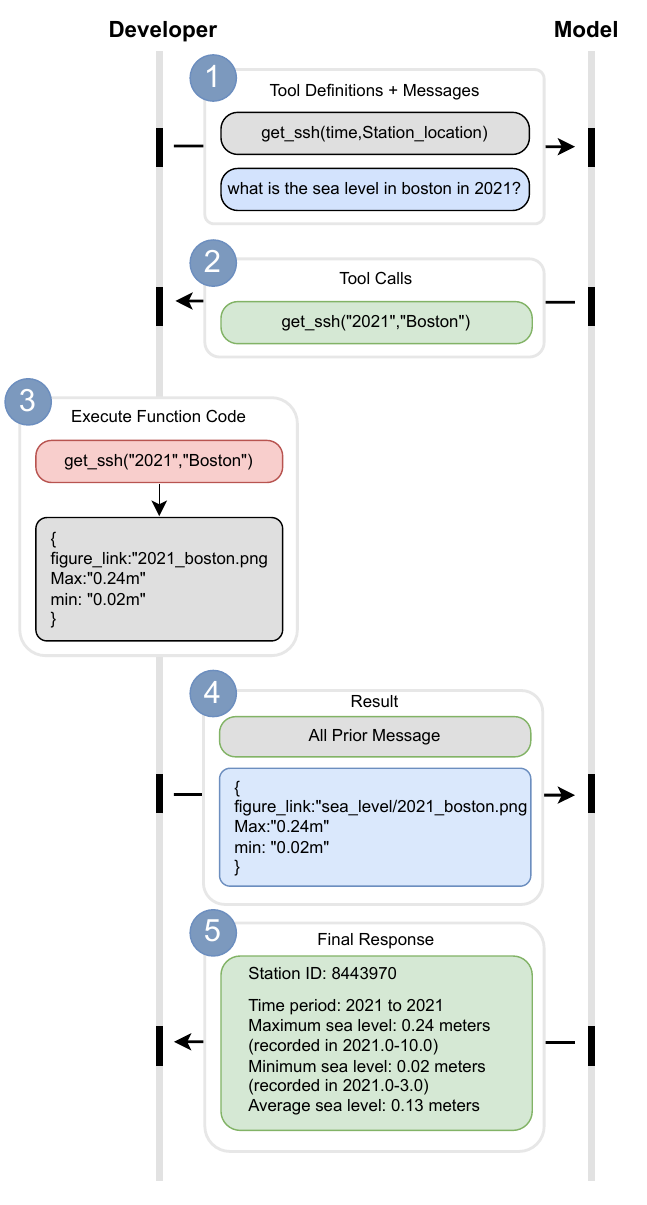}
 \caption{%
 Function-calling workflow in large language models (LLMs). 
The LLM parses a user query (e.g., \textit{What is the sea level in Boston in 2021?}) 
into a structured API call (e.g., \texttt{get\_ssh(2021, Boston)}). 
The call is executed externally, and the returned results 
(e.g., figures or statistics) are combined into the final response~\citep{openai_function_call}.}
\label{fig:function-calling}
\end{figure}
\textbf{Function Calling.}
A key mechanism enabling agentic behavior is \textbf{function calling}, where the model emits structured outputs (typically in JSON\footnote{A lightweight, text-based format for structured data exchange. JSON (JavaScript Object Notation) is commonly used to encode hierarchical data such as parameter sets, metadata, and API calls, and is well suited for integrating machine learning models with scientific tools and datasets.} format) to request downstream tool execution~\citep{schick2023toolformer,openai_function_call}.As illustrated in \textbf{Figure~\ref{fig:function-calling}}, this process follows a loop of interpretation, execution, and synthesis: a user query such as “What is the sea level in Boston in 2021?” is first translated by the model into a structured API call (e.g., \texttt{get\_ssh("2021", "Boston")}). The external function is then executed—retrieving sea surface height (SSH) data—and the result (e.g., image links and numerical statistics) is returned to the model. The LLM integrates this structured output into a natural language response that is both accurate and verifiable.
This approach enables LLMs to act as dynamic interfaces to computational infrastructures, making it possible to construct interactive and trustworthy AI systems for scientific and engineering domains. By incorporating real-time function execution and grounding model outputs in factual data, function calling enhances both the transparency and utility of LLM-driven workflows.

\section{Related Work}
\label{sec:related_work}
\vspace{-5pt}
\textbf{LLMs in Scientific Domains.} A central challenge in applying large language models (LLMs) to scientific domains lies in ensuring factual accuracy, reproducibility, and timeliness~\citep{ji2023survey}. Although LLMs excel in language fluency and general reasoning, their knowledge is static and often opaque, making it difficult to verify or update information. This limitation is critical for scientific applications, where outputs must be grounded in authoritative, up-to-date datasets and follow domain-specific standards. In climate and Earth sciences, for example, models that cannot handle structured formats such as NetCDF or GRIB risk misinterpretation of quantitative results. Recent work in biomedical~\citep{singhal2023large} and materials science~\citep{mostafa2024grag} has similarly shown that domain adaptation is essential for reliable use in high-stakes research contexts.

\textbf{Tool Augmentation and Retrieval-Augmented Generation.} A widely studied mitigation strategy is \textit{tool augmentation}, where LLMs invoke external tools for information retrieval or computation~\citep{schick2023toolformer}. Toolformer and related agent-based systems such as HuggingGPT~\citep{shen2023hugginggpt} demonstrate that models can be trained to autonomously call APIs such as search engines, calculators, or code execution environments, improving factuality. Retrieval-augmented generation (RAG) further enhances this by grounding responses in retrieved documents, as seen in WebGPT~\citep{nakano2022webgpt}. However, these frameworks primarily operate on unstructured text and lack native support for specialized scientific data formats or spatiotemporal datasets.

\textbf{LLMs for Earth and Ocean Sciences.} In the Earth sciences, LLM integration is still nascent. Domain-specific models such as \textit{ClimateGPT}~\citep{thulke2024climategpt} and \textit{GeoGalactica}~\citep{lin2024geoga} have shown promise in processing literature and reports, but they do not incorporate real-time observational data streams. In the ocean domain, \textit{OceanGPT}~\citep{bi2024oceangpt} fine-tunes large language models on a multi-agent–generated instruction dataset and introduces OCEANBENCH for benchmarking 15 ocean science tasks. While it improves domain-specific reasoning compared with general LLMs, it operates primarily on static corpora and lacks real-time integration with authoritative observational datasets, limiting its applicability for time-sensitive reporting.A recent study further introduced the Intelligent Data Exploring Assistant (IDEA), which integrates LLMs with domain-specific geoscience data and analytical tools~\citep{Widlansky2025}.

\textbf{The Novelty of Our Method.} \textbf{OceanAI} addresses these limitations by combining natural language interfaces with a domain-specialized function-calling architecture that retrieves, preprocesses, and visualizes live data from trusted providers such as NOAA. This design ensures outputs are not only contextually accurate but also up-to-date, transparent, and fully reproducible, making it suitable for operational coastal and oceanographic monitoring.

\section{System Design Of OceanAI}
\begin{figure}[!t]
  \centering
  \includegraphics[width=0.95\textwidth]{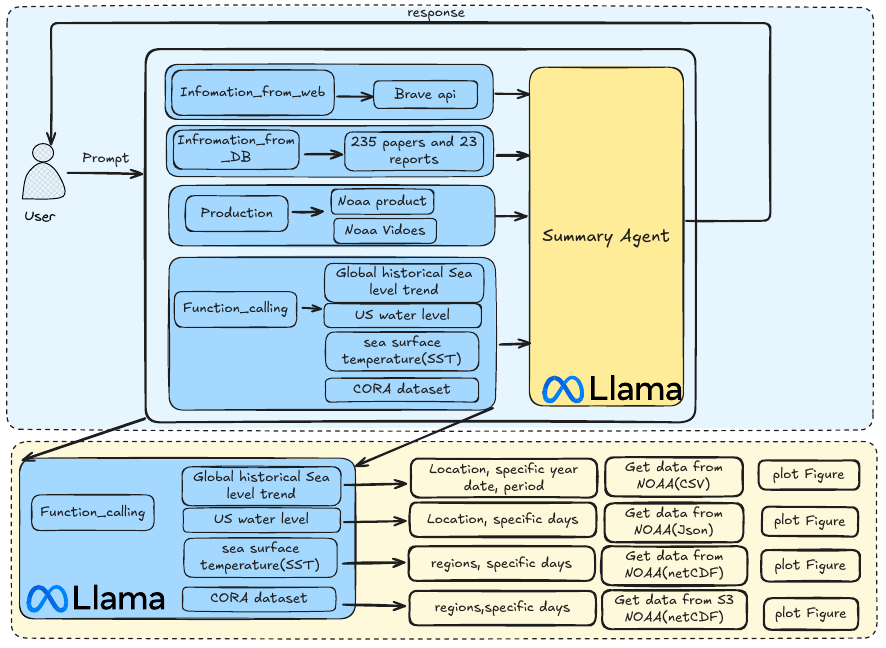}
  \caption{\textbf{OceanAI system architecture.} The unified agent-based pipeline executes fixed modules for web retrieval (via Brave API), document search (via Chroma), and curated NOAA media. Function calling is dynamically triggered based on query semantics, providing structured access to NOAA datasets such as SST, sea level, and CORA via parameterized back-end functions.}
  \label{fig:system_architecture}
\end{figure}
\subsection{Architecture Overview}
OceanAI is implemented as a multi-agent system based on tool-augmented large language models (LLMs), where a single coordinating agent orchestrates the execution of a fixed pipeline composed of multiple domain-specific modules. Each module is responsible for interfacing with a different category of information source, and together they enable comprehensive, multi-modal responses to user queries.

As illustrated in \textbf{Figure~\ref{fig:system_architecture}}, when a user submits a natural language prompt, the system routes the request to a centralized LLaMA-based agent augmented with function calling capabilities. This agent subsequently executes a sequence of predefined modules to gather relevant information. Specifically, the architecture includes the following components:

\begin{itemize}
    \item \textbf{Web Retrieval Module}: Uses the Brave Search API to obtain up-to-date information from the internet. This step is always executed to supplement the response with recent developments or external references.
    \item \textbf{Document Retrieval Module}: Connects to a Chroma vector database containing embeddings of 235 scientific papers and 23 NOAA technical reports. Retrieved passages are selected based on similarity to the query and used for evidence-based summarization.
    \item \textbf{Media Production Module}: Returns curated NOAA-produced content, such as multimedia videos and official bulletins, which provide reliable public-facing context.
    \item \textbf{Structured Data Module}: This is the only dynamic component in the system. When the user request involves geospatial or numerical data (e.g., temperature, sea level), the agent uses function calling to invoke one or more back-end functions. These functions retrieve and process data from NOAA datasets in various formats (CSV, JSON, NetCDF) and return results in structured or visual form. The detailed structure and extensibility of this function calling framework are described in Section~\ref{sec:implementation}.

\end{itemize}
All retrieved information—whether textual or numerical—is aggregated and summarized by the central agent before returning a final response to the user. The architectural design supports future expansion through the registration of additional callable functions and modular data sources.

\subsection{Modular Function-Calling System}
\label{sec:func-calling}
OceanAI uses a modular \emph{function-calling} layer that maps natural-language prompts to typed, parameterized analysis functions. Each function targets a specific oceanographic task (e.g., water level, sea level trends, CORA reanalysis, SST) with arguments for space, time, and variable selection. A dispatcher infers parameters from the prompt, invokes the appropriate function, and returns results grounded in authoritative sources (NOAA CO-OPS, CORA, CRW) spanning real-time APIs, gridded reanalyses, and long-term observational archives (see \textbf{Table~\ref{tab:datasets}}).

\begin{table}[!t]
  \centering
  \includegraphics[width=\textwidth]{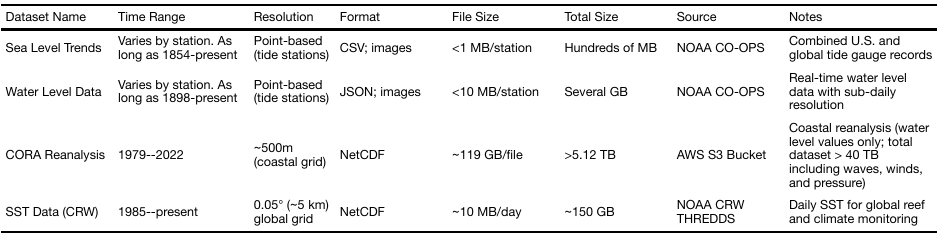}
  \caption{Datasets accessible through OceanAI’s function-calling interface, including real-time gauge measurements and historical/gridded products from NOAA or AWS-hosted repositories.}
  \label{tab:datasets}
\end{table}

\begin{figure}[!thb]
  \centering
\includegraphics[width=0.45\linewidth]{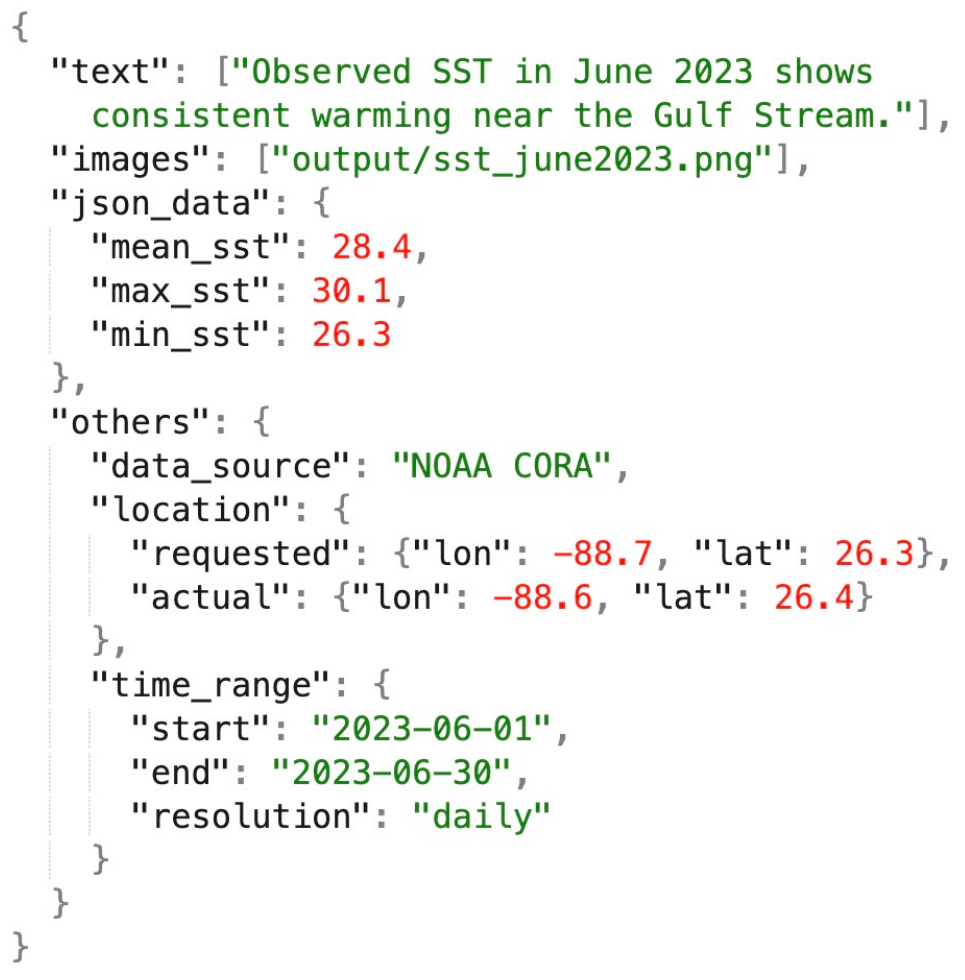}
  \caption{Standardized function output schema in OceanAI, including \texttt{text}, \texttt{images}, \texttt{\detokenize{json_data}}, and \texttt{others} (metadata: location, units, time coverage, source).}
  \label{fig:json_output}
\end{figure}

To ensure consistency and downstream integration, every function returns a standardized payload with four fields: a natural-language summary (\texttt{text}), visualization paths (\texttt{images}), structured statistics/series (\texttt{\detokenize{json_data}}), and metadata (\texttt{others}) covering source, units, spatial context, and time span (example in \textbf{Figure~\ref{fig:json_output}}). This separation of language understanding from data access/transformation improves interpretability, reproducibility, and extensibility; adding a new dataset or routine only requires registering a new function.

\subsection{System Implementation}
\label{sec:implementation}
\begin{table}[!htb]
\renewcommand{\arraystretch}{1.1} 
\small
\centering
\begin{tabular}{lp{0.22\linewidth}p{0.58\linewidth}}
\hline
\textbf{Layer} & \textbf{Technology} & \textbf{Description} \\
\hline
Frontend & Next.js & React-based framework supporting SSR and static site generation. Provides an interactive UI for user queries and results. \\
Backend & Flask & Lightweight Python web server for handling requests, APIs, and model orchestration. \\
LLM Model & meta-llama/llama-4-scout-17b-16e-instruct & Instruction-tuned LLaMA 4 model supporting advanced NL understanding and generation. \\
Database & ChromaDB & Embedding-aware vector DB supporting semantic retrieval and RAG workflows. \\
Deployment & AWS EC2 & Cloud-based VM hosting backend, frontend, and LLM runtime. Oceanographic data is accessed via S3. \\
\hline
\end{tabular}
\caption{\label{techstack}Summary of technologies used in OceanAI’s implementation. Each layer handles specific tasks from user interface rendering to semantic retrieval and cloud deployment.}\vspace{-1.5em}
\end{table}
OceanAI is implemented using a modular and lightweight technology stack designed for interactive scientific reasoning. The system architecture consists of a Next.js frontend for user interaction and visualization, a Flask-based backend responsible for managing API requests and orchestrating function calls, and a large language model (meta-llama/llama-4-scout-17b-16e-instruct) deployed with instruction-following capabilities.

For data retrieval and document grounding, the system uses ChromaDB, an embedding-based vector database that supports dense retrieval for retrieval-augmented generation. All services are containerized and deployed on AWS EC2 instances, with large-scale environmental datasets accessed through S3-based storage or NOAA-hosted endpoints. \textbf{Table~\ref{techstack}} summarizes the major components of the implementation stack.

\section{User Case Demonstration}
\label{sec:usercases}

To illustrate the practical capabilities of \textbf{OceanAI}, we present four representative use cases, each targeting a distinct data modality and query type. In each case, a natural language prompt triggers a corresponding function call that retrieves, processes, and visualizes structured scientific data. All outputs are grounded in authoritative NOAA datasets, including CO-OPS sea level archives, CORA reanalysis, and Coral Reef Watch (CRW) SST products.

\subsection{Sea Level Trends from CSV}
\label{sec:case-sea-level}

\textbf{Prompt:} \textit{“What is the sea level in Boston and Virginia Key in 2022?”}

\begin{figure}[!thb]
    \centering
    \fbox{\includegraphics[width=0.95\textwidth]{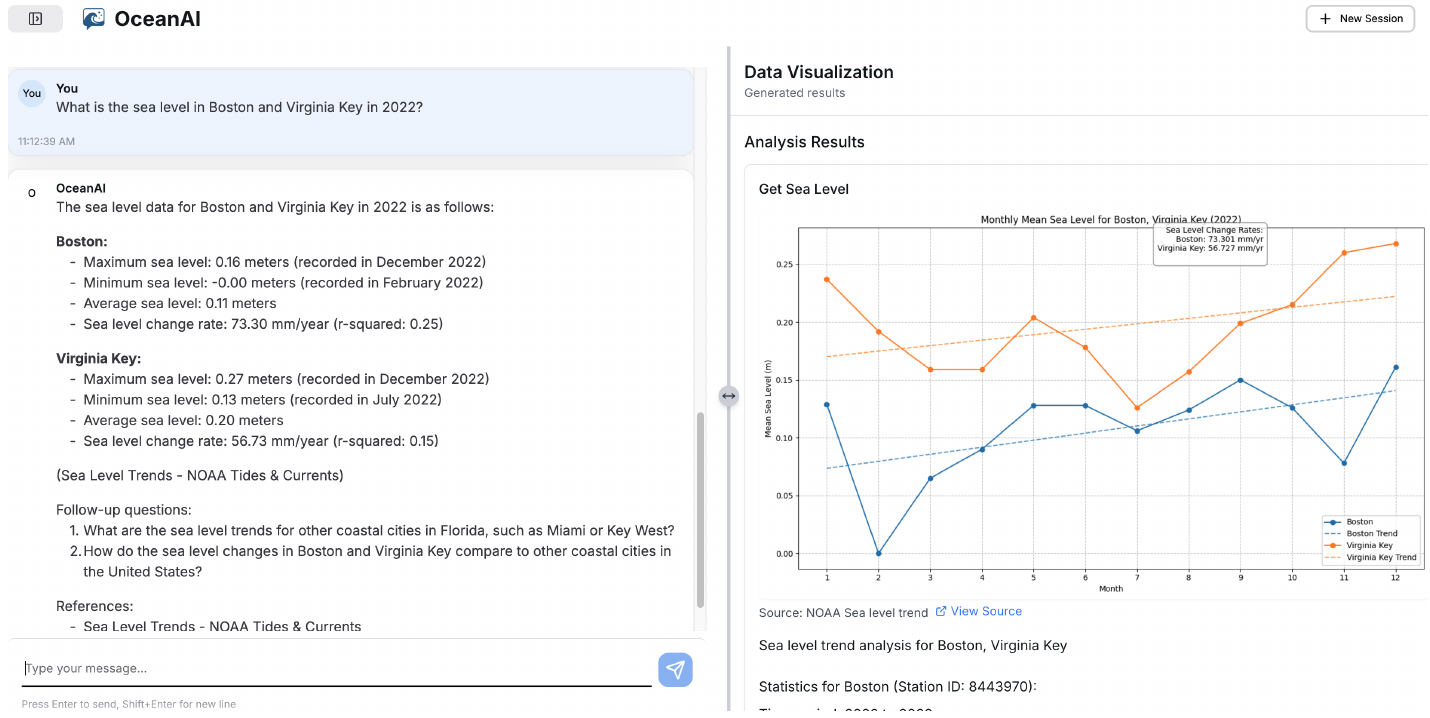}}
    \caption{OceanAI output for a sea level trend query comparing Boston and Virginia Key in 2022. Data from NOAA CO-OPS includes monthly mean sea levels, annual change rates, and plotted trends for both locations.}\vspace{0.5em}
    \label{fig:case-sea-level}
\end{figure}

OceanAI retrieves historical sea level records in CSV format from NOAA CO-OPS, selects the appropriate tide stations, and returns monthly mean values along with calculated change rates. Results include a comparative time series plot for both locations.(\textbf{Figure~\ref{fig:case-sea-level}}).
\vspace{-5pt}

\subsection{Water Level Monitoring from JSON}
\label{sec:case-water-level}

\textbf{Prompt:} \textit{“What was the water level in Boston in May 2020?”} 

\begin{figure}[!htb]
    \centering    \fbox{\includegraphics[width=0.95\textwidth]{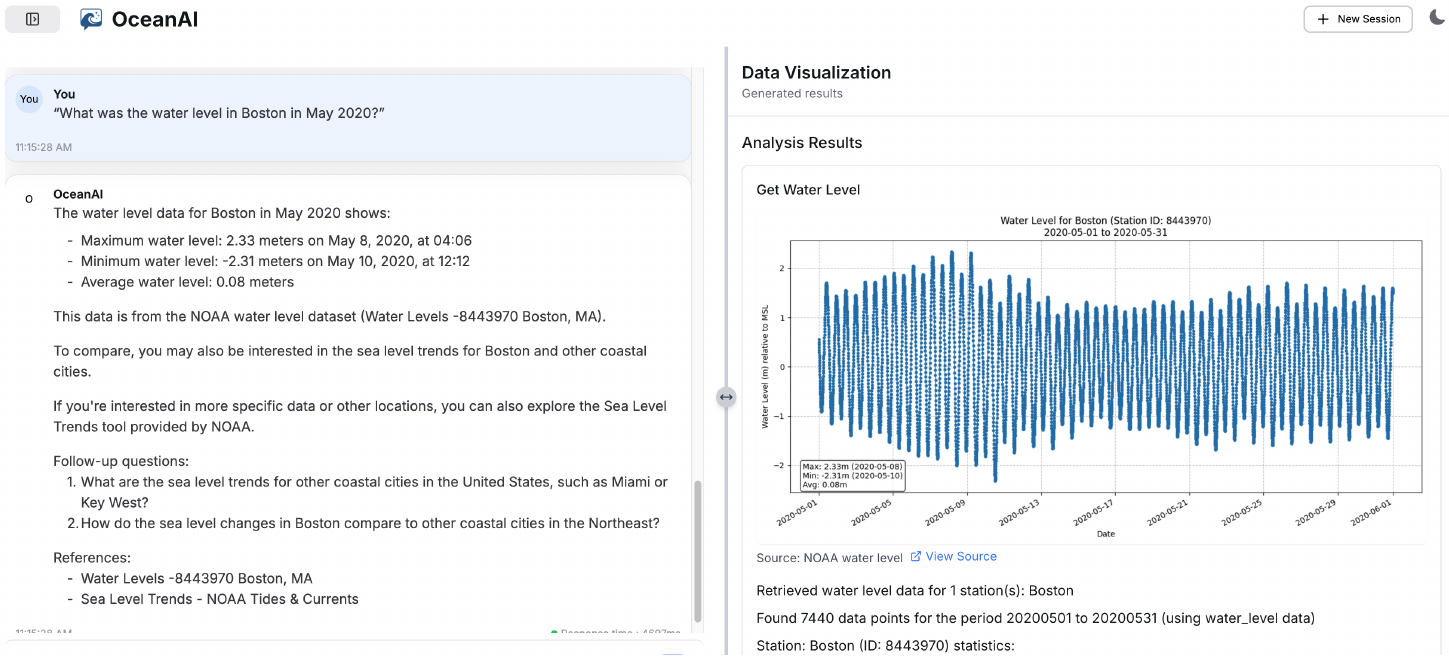}}
    \caption{OceanAI output for a real-time water level query in Boston for May 2020. NOAA CO-OPS data is visualized as an hourly tidal time series with annotated maximum, minimum, and average values.}\vspace{-1.5em}
    \label{fig:case-water-level}
\end{figure}

This query accesses high-frequency JSON records from NOAA's CO-OPS tide gauge network. OceanAI extracts hourly water level measurements for the specified date range and produces a high-resolution tidal time series plot annotated with extrema, timestamps, and station metadata (\textbf{Figure~\ref{fig:case-water-level}}).

\subsection{CORA Reanalysis Profile (NetCDF)}
\label{sec:case-cora}

\textbf{Prompt:} \textit{“Show Boston's water level from CORA reanalysis in June 1993.”} 

\begin{figure}[!htb]
    \centering
    \fbox{\includegraphics[width=0.95\textwidth]{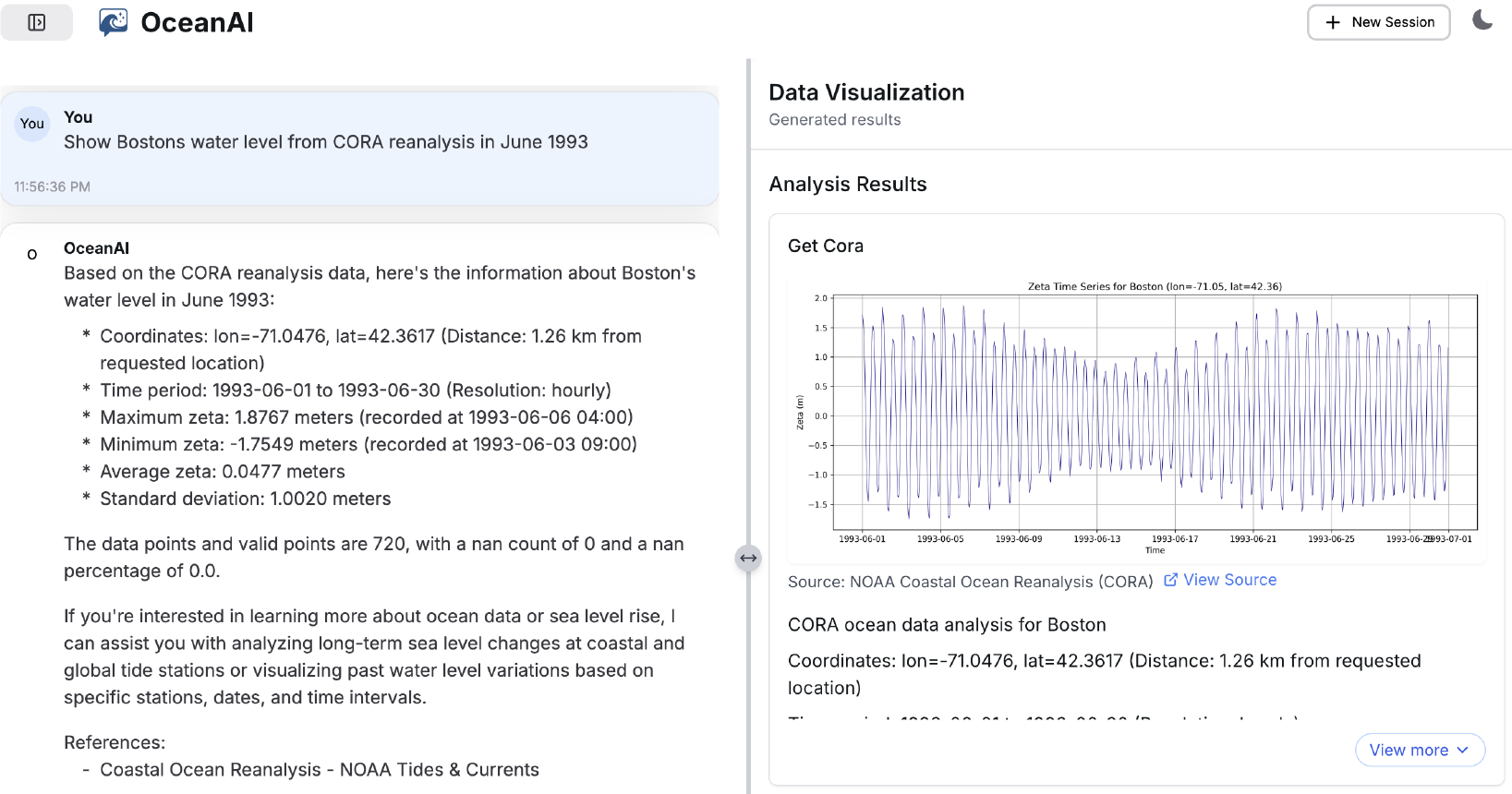}}
    \caption{OceanAI output from the NOAA CORA reanalysis showing Boston’s water level (zeta) for June 1993, including maximum, minimum, average, and standard deviation values with corresponding time series plot.}\vspace{-1.5em}
    \label{fig:case-cora}
\end{figure}

OceanAI queries the NOAA Coastal Ocean Reanalysis (CORA) dataset in NetCDF format, selecting grid points based on spatial proximity and filtering by time and variable. The resulting time series plot illustrates water level (zeta) variation over the specified two-month period, along with computed statistics (\textbf{Figure~\ref{fig:case-cora}}).

\subsection{Sea Surface Temperature from Gridded SST}
\label{sec:case-sst}

\textbf{Prompt:} \textit{“What was the SST in the Gulf of Mexico in 2019?”} 

\begin{figure}[htb]
    \centering
    \fbox{\includegraphics[width=0.95\textwidth]{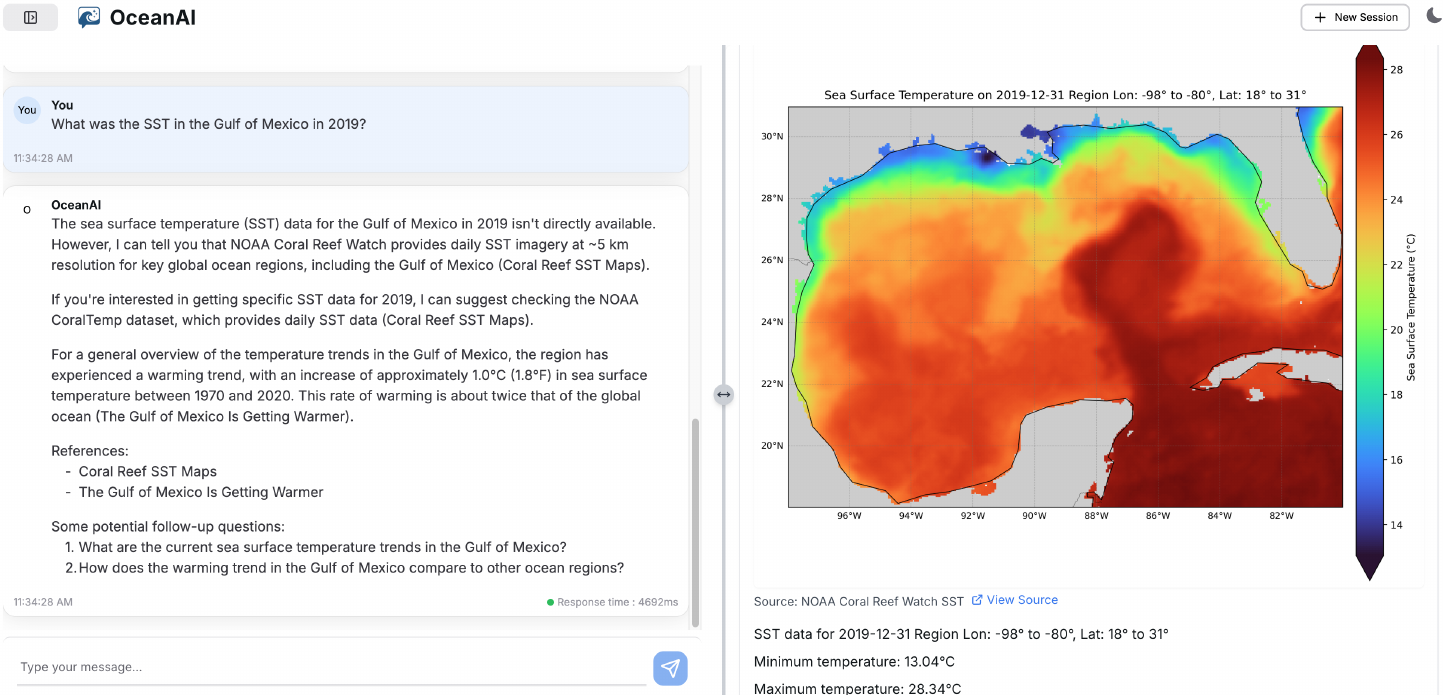}}
    \caption{OceanAI output retrieving a sea surface temperature (SST) map from the NOAA Coral Reef Watch dataset for the Gulf of Mexico on December 31, 2019. The map shows spatial temperature distribution with recorded extremes of 13.04 degrees Celsius and 28.34 degrees Celsius.}\vspace{-1em}
    \label{fig:case-sst}
\end{figure}

This query retrieves NOAA Coral Reef Watch (CRW) SST data in NetCDF format, filters for the Gulf of Mexico in the specified year, and visualizes the spatial temperature distribution. Results include minimum and maximum SST values for the selected date and region (\textbf{Figure~\ref{fig:case-sst}}).

\section{Comparison}

\subsection{Existing Related Platforms and Tools}

\begin{table}[!htb]
\centering
\includegraphics[width=\textwidth]{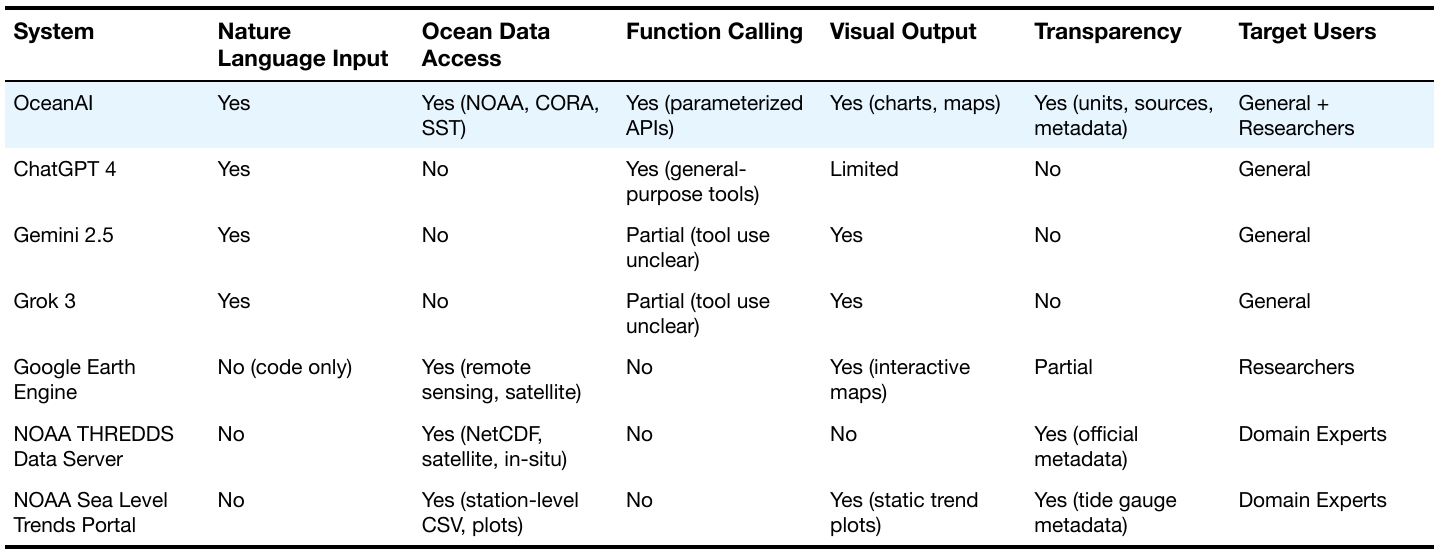}
\caption{Comparison of \textbf{OceanAI} with general LLMs, geospatial platforms, and NOAA portals across interface, data access, function calling, visualization, transparency, and target users. OceanAI uniquely integrates conversational queries with parameterized access to authoritative NOAA datasets, producing chart- and map-based outputs with full metadata.}\vspace{-0.1em}
\label{tab:related_platforms}
\end{table}

To contextualize OceanAI's functionality, we compare it with a range of existing platforms and tools commonly used for oceanographic data access, interaction, and analysis (\textbf{Table~\ref{tab:related_platforms}}). These systems span general-purpose conversational agents, domain-specific data portals, and geospatial computation frameworks.

General LLM-based platforms such as ChatGPT and Gemini 2.5 support natural language interaction and, in some configurations, basic tool integration. However, they lack direct access to authoritative ocean datasets and typically cannot execute parameterized scientific queries. While they may produce fluent explanations, their outputs are often ungrounded and unverifiable, limiting their use in data-driven scientific workflows.

On the other end of the spectrum, domain-specific platforms like the NOAA THREDDS Data Server and the NOAA Sea Level Trends Portal provide structured, high-quality observational data. These tools support transparent access to metadata and historical records but are designed for expert users. They do not offer natural language interfaces or flexible data query mechanisms. Users are expected to navigate file hierarchies, understand specialized formats (e.g., NetCDF, CSV), and perform post-processing independently.

Google Earth Engine provides an advanced, scalable platform for satellite-based geospatial analysis. While it supports global ocean data layers and visualization, it requires users to write JavaScript or Python code. Furthermore, it lacks support for conversational interaction and real-time function invocation over custom parameter sets.

In contrast, OceanAI uniquely integrates natural language interfaces with structured, callable scientific data functions. It supports verifiable outputs, multimodal responses (text, figures, and structured data), and extensibility through modular function registration. As such, OceanAI bridges the usability gap between general-purpose chatbots and expert-level data portals, enabling interdisciplinary access to ocean science data with both transparency and technical rigor.






\subsection{Empirical Comparison}
\vspace{-5pt}
We evaluated OceanAI's performance through a series of experiments designed to test its three core capabilities: 
retrieving up-to-date information from the web, accessing embedded scientific reports, and analyzing structured ocean data from NOAA datasets. 
The results were compared with three AI chat-interface products: ChatGPT (OpenAI), Gemini (Google), and Grok (xAI).

Each task was categorized as either unstructured or structured. 
Unstructured tasks evaluated how well each system could retrieve and summarize information from NOAA web pages and embedded reports. 
Structured tasks involved direct access to authoritative NOAA datasets for data retrieval and analysis.

\begin{table}[!htb]
\centering
\includegraphics[width=\linewidth]{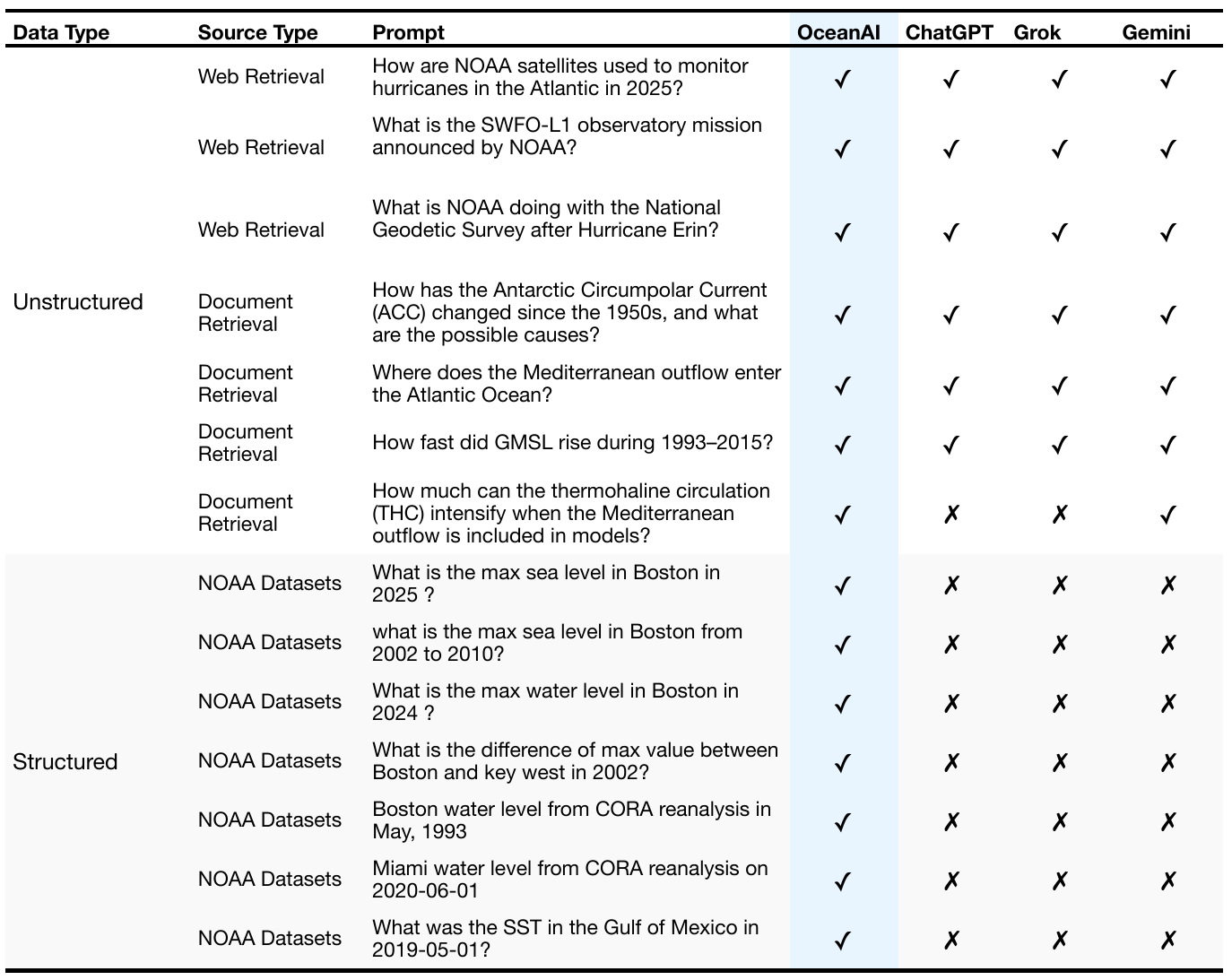}
\caption{\textbf{Comparison of OceanAI and AI chat-interface products on representative ocean science tasks.}
A checkmark (\ding{51}) indicates a correct and verifiable answer, while a cross (\ding{55}) indicates an incorrect or incomplete response.}
\label{tab:ai-extended-comparison}
\end{table}

As summarized in \textbf{{Table~\ref{tab:ai-extended-comparison}}},OceanAI consistently produced accurate and traceable results across both unstructured and structured tasks. 
In contrast, ChatGPT, Gemini, and Grok performed reasonably well on unstructured information retrieval tasks 
but failed to access or analyze structured NOAA data through dataset queries. 
These results demonstrate OceanAI’s advantage in integrating real-time data access with domain-specific reasoning and transparent source attribution.Additional qualitative outputs and partial example screenshots corresponding to selected evaluation cases are provided in Appendix A.

\subsection{Theoretical Analysis}

The experimental results show that OceanAI performs better than AI chat-interface products in both unstructured and structured ocean data tasks. 
Its tool-augmented architecture separates the LLM’s role in language understanding from external modules that perform data retrieval and computation, 
ensuring that outputs are derived from verified data sources rather than model priors. 
This design improves transparency and reproducibility in scientific applications.

\textbf{Web and document retrieval.}
OceanAI integrates real-time web access and embedded literature retrieval to obtain up-to-date and verifiable NOAA information. 
This enables accurate responses to unstructured queries by reducing hallucination and maintaining factual consistency across both current and historical sources.

\textbf{Structured data module.}
For structured numerical tasks, OceanAI uses a function-calling framework that directly queries NOAA datasets through defined parameters such as region, time, and variable. 
This ensures that numerical results are based on authoritative data with traceable provenance and controlled accuracy, explaining its superior performance on quantitative tasks.

\textbf{System design.}
Although not directly tested in this study, OceanAI’s modular structure allows new data sources and analytical routines to be added without retraining the model. 
This supports adaptability and maintainability as ocean data systems evolve.

Overall, OceanAI links natural-language reasoning with verifiable data execution, enabling accurate, transparent, and domain-grounded outputs in the ocean domain.
\section{Conclusion}

We presented OceanAI, a modular framework that augments large language models with tools to interact directly with real-world oceanographic datasets through natural language. OceanAI addresses core challenges in applying LLMs to scientific domains, including hallucination, lack of transparency, and difficulty in handling structured formats such as NetCDF and JSON. Through representative use cases on sea level observations, water level reanalysis, and sea surface temperature records, we demonstrated how the system produces interpretable and verifiable outputs. The framework also incorporates document-based reasoning via vector search, enabling the integration of structured data with contextual understanding.

The extensible design of OceanAI supports the inclusion of additional datasets, analytical tools, and multi-agent architectures to advance collaborative scientific reasoning. Future work will focus on incorporating uncertainty quantification, multimodal analysis, and broader environmental data sources. These developments aim to establish OceanAI as a general-purpose assistant for transparent and trustworthy environmental intelligence.
\section*{Acknowledgements}


Research support provided through NSF grant OCE-1559178, OCE-1851421, RISE-2019758, OCE-2206052, CNS-2223844, OAC-2417850, OAC-2514351, and SECOORA grant NA16NOS0120028 is much appreciated. We thank Mr. Derrick Snowden for insightful discussions and support throughout this study, and Dr. John Callahan for his assistance in compiling the database of relevant publications.

\bibliographystyle{plainnat}
\bibliography{latex/main}

\clearpage
\appendix
\renewcommand\thesection{Appendix:}
\renewcommand\thesubsection{\arabic{subsection}}
\renewcommand{\thefigure}{\Alph{section}\arabic{figure}}
\setcounter{figure}{0}

\section{Model Comparison on Structured and Unstructured Data}

This appendix presents comparative outputs from OceanAI and general-purpose large language models
(ChatGPT, Gemini, and Grok) for both structured and unstructured marine-science tasks.
OceanAI consistently provides domain-grounded, source-linked, and numerically accurate results,
while baseline models offer descriptive but less verifiable responses.

\subsection{Structured Data Comparison}

\begin{figure}[h]
    \centering
    \includegraphics[width=0.95\textwidth]{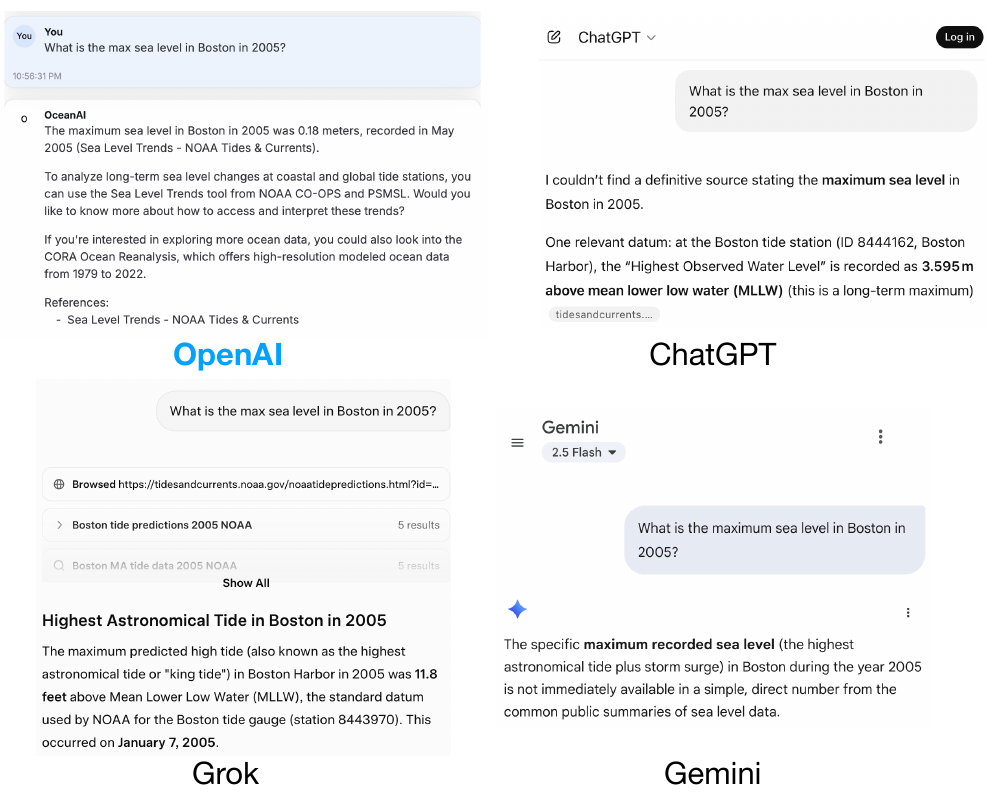}
    \caption{Comparison of model responses to a structured NOAA query:
    “What is the maximum sea level in Boston in 2005?”
    OceanAI retrieves an exact numeric value with source attribution,
    while ChatGPT, Gemini, and Grok provide incomplete or text-only answers.}
\end{figure}

\begin{figure}[h]
    \centering
    \includegraphics[width=0.95\textwidth]{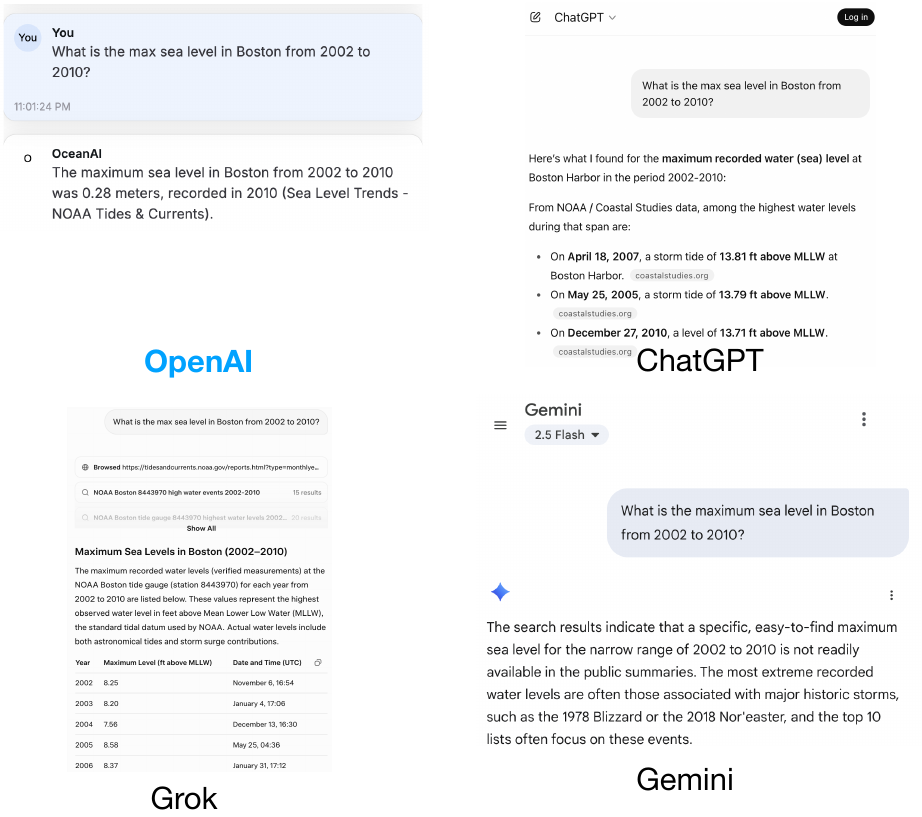}
    \caption{Comparison of model responses to a structured query across multiple years:
    “What is the maximum sea level in Boston from 2002 to 2010?”
    OceanAI extracts and formats verified NOAA records.}
\end{figure}

\clearpage
\subsection{Unstructured Data Comparison}

\begin{figure}[h]
    \centering
    \includegraphics[width=0.95\textwidth]{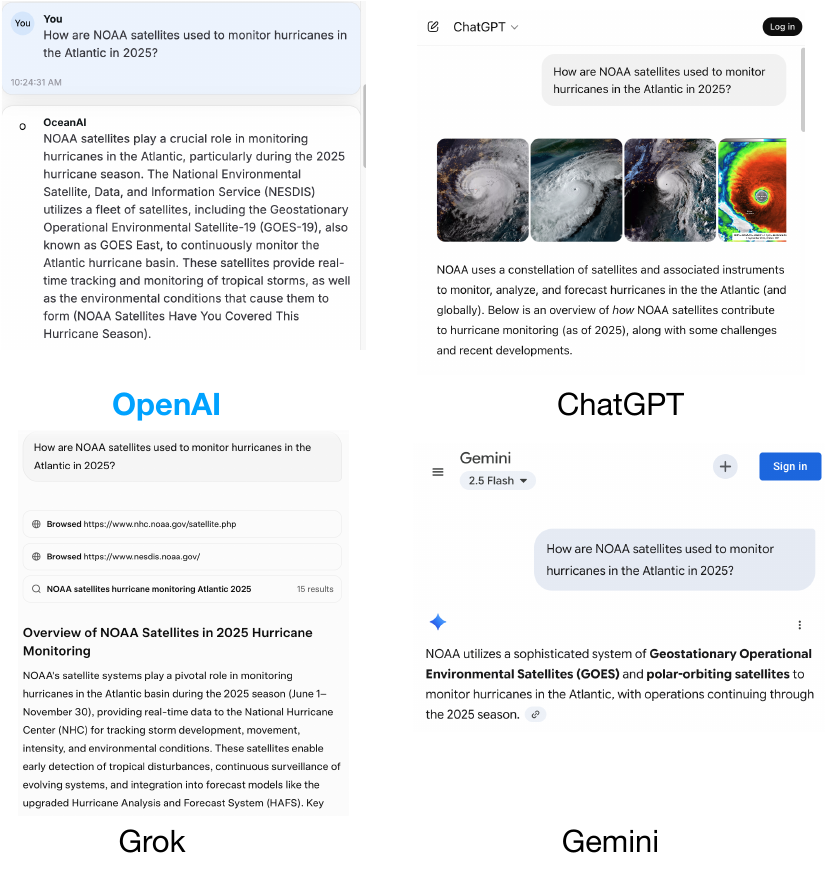}
    \caption{Model responses to an unstructured question:
    “How are NOAA satellites used to monitor hurricanes in the Atlantic in 2025?”}
\end{figure}

\begin{figure}[h]
    \centering
    \includegraphics[width=0.95\textwidth]{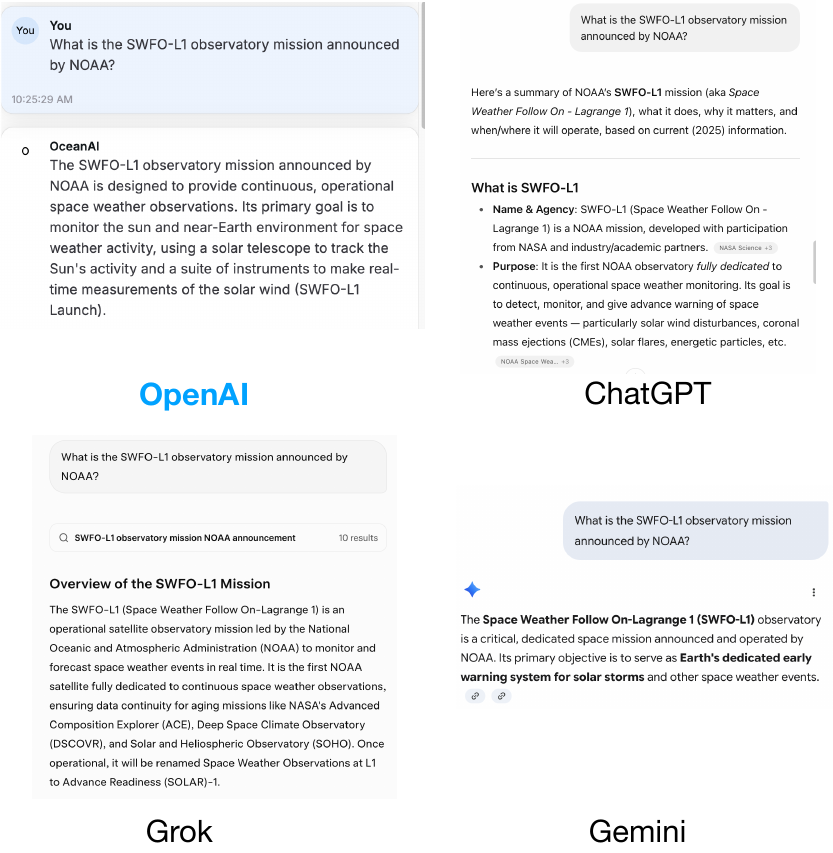}
    \caption{Model responses to an unstructured scientific query:
    “What is the SWFO-L1 observatory mission announced by NOAA?”}
\end{figure}

\end{document}